\documentclass[conference]{IEEEtran}
\IEEEoverridecommandlockouts
\usepackage{cite}
\usepackage{amsmath,amssymb,amsfonts}
\usepackage{algorithmic}
\usepackage{graphicx}
\usepackage{subfigure}
\usepackage{textcomp}
\usepackage{xcolor}
\usepackage{hyperref}
\usepackage{caption}
\usepackage{multirow}
\def\BibTeX{{\rm B\kern-.05em{\sc i\kern-.025em b}\kern-.08em
    T\kern-.1667em\lower.7ex\hbox{E}\kern-.125emX}}
\begin{document}

\title{Forest and Water Bodies Segmentation Through Satellite Images Using U-Net}

\author{\IEEEauthorblockN{Dmytro Filatov}
\IEEEauthorblockA{\textit{Artificial Intelligence and Computer Vision} \\
\textit{Aimech Technologies Corp.}\\
San Francisco, The United States of America \\
dmytro@aimechtechnologies.com}
\and
\IEEEauthorblockN{Ghulam Nabi Ahmad Hassan Yar}
\IEEEauthorblockA{\textit{Department of Electrical and Computer Engineering} \\
\textit{Air University}\\
Islamabad, Pakistan \\
gnahy1@gmail.com}
}

\maketitle

\begin{abstract}
Global environment monitoring is a task that needs attention due to changing climate. This includes monitoring the rate of deforestation and areas affected by flooding. Satellite imaging has helped a lot in effectively monitoring the earth, and deep learning techniques have helped automate this monitoring process. This paper proposes a solution for observing the area covered by the forest and water. To achieve this task UNet model has been proposed, which is an image segmentation model. Our model achieved a validation accuracy of 82.55\% and 82.92\% for segmentation of areas covered by forest and water, respectively.
\end{abstract}


\section{Introduction}

The development of technology has helped solve many problems and has improved the status of living in the modern age. Deep learning is one of the significant advancements in the latest technologies, and it has helped in automating many tasks. Due to this automation, many tasks can now be done in a matter of minutes. Neural networks (NN) have shown an essential part in solving modern-day problems related to computer vision. It has shown its importance in image classification, object detection, and segmentation. Currently, neural networks have shown their worth in solving almost all the problems related to computer vision \cite{goodfellow2016deep}.

Convolutional Neural Networks (CNN) have helped ease the process of achieving many applications in the field of computer vision. In classical machine learning models, i.e., support vector machine (SVM), random forest, decision tree (DT), etc., are used for classification. For computer vision applications to implement classification, these models need extracted features. These features are extracted through various filters, i.e., the bag of words model. A convolutional neural network is a complete package that has convolutional layers as feature extractors and fully connected layers as a classifier. Convolutional layers have pooling layers and Dense layers in between them. These pooling layers in between help in the downsampling of the input image. The main task of convolutional layers is to extract the features from the images. Fully connected layers act as classification layers. Features extracted through convolutional layers are flattened and fed to fully connected layers for classification. Fully connected layers can also be replaced by classical machine learning models mentioned above, as features are extracted through convolutional layers.

The problem of satellite image segmentation is challenging and can also be solved using neural networks \cite{khryashchev2018comparison}. Satellite images can help in overviewing the landscapes, and weather, and help in monitoring the changes that happen over time. This includes monitoring the direction of storms by following the eye of the cyclon, trends of forestation of deforestation, and drying of lakes or rivers. To carry out any of these tasks, first, the satellite images need to be segmented, i.e., segmenting the eye of a cyclone, segmenting the trees, or water bodies is necessary to carry out the applications mentioned above.

U-Net is a type of convolutional neural network used for image segmentation that has been widely used in satellite image segmentation. U-Net is vastly used because of its properties like fast and precise segmentation of images \cite{ronneberger2015unet}. In this network, at first, the features are extracted through downsampling, and on the basis of these features, segmented masks of the image are generated via upsampling. As this is a supervised model, it needs data to be trained. During training, these networks take images and their corresponding masks to learn the features that lie inside the mask so that these features can be segmented during real-world testing.

\begin{figure*}
    \centering
    \includegraphics[scale=0.3]{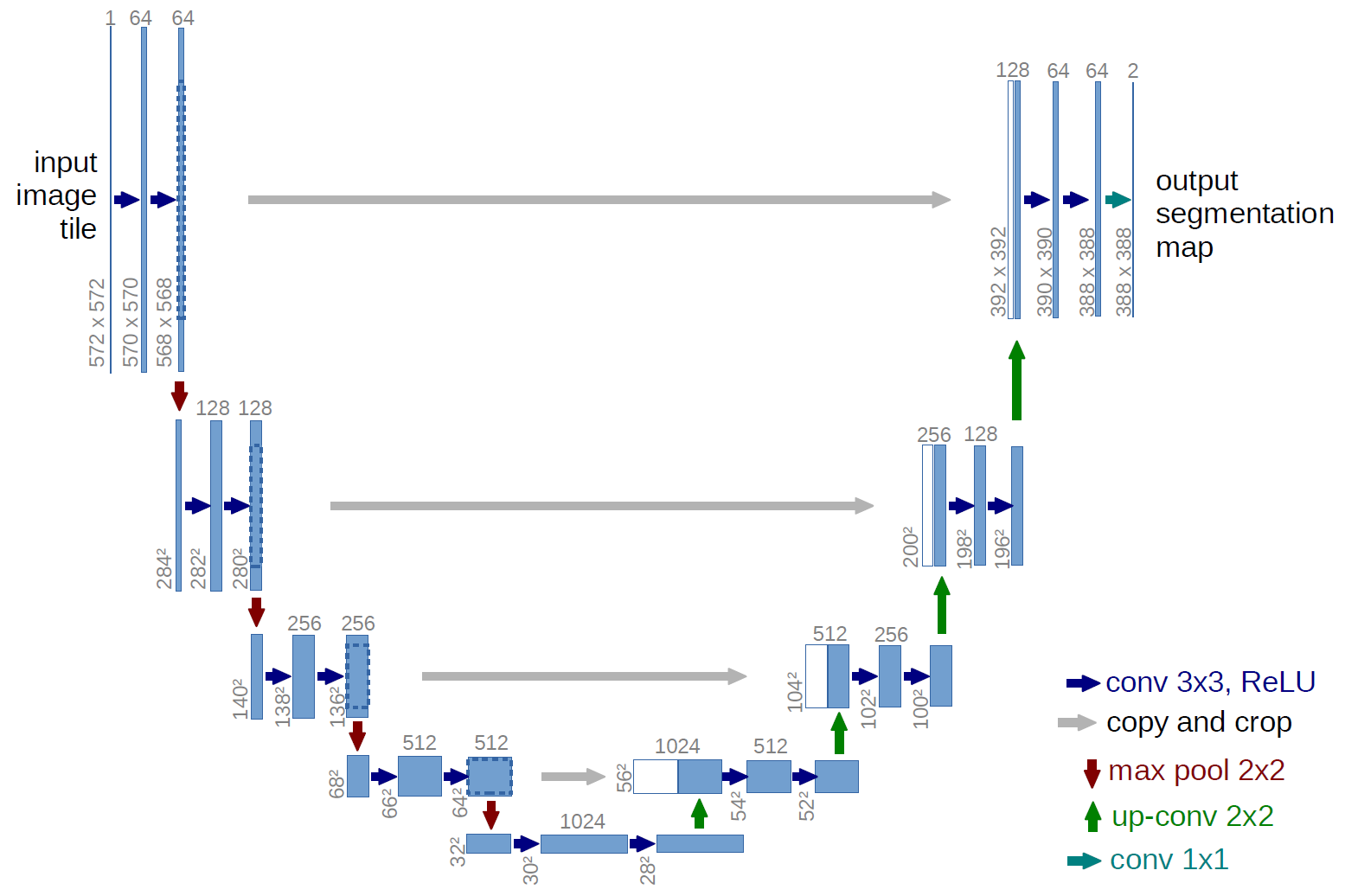}
    \caption{Basic architecture of U-Net \cite{ronneberger2015unet}}
    \label{fig:unet}
\end{figure*}

This paper proposes the application of forest and water bodies segmentation in satellite images using the U-Net model. This segmentation can help in monitoring the forestation, deforestation, flooded areas, and drying of lakes or rivers over time. The goals achieved by this paper are:

\begin{itemize}
    \item Segmentation of forests
    \item Segmentation of water bodies
\end{itemize}

The rest of the paper is divided as follows: Section 2 presents the work that has already been done in the field of segmentation. Section 3 presents the research methodology along with the system specifications and model used. This section also highlights the implementation parameters of the model. Section 4 is about the dataset. This section explains the sources of the dataset and its properties. This section also shows the sample images and their respective masks for the datasets. Section 5 will be about the results and discussion on them. This section shows the accuracy and loss curves for both training and validation datasets. Results will also be discussed in this section. A review on the basis of the output curved will be given in this section. In the end, the paper will be concluded with final comments, and future work that can improve the experiment or overall application will be presented. After that, references used in this study will be given.

\begin{figure*}
         \centering
         \includegraphics[scale = 0.5]{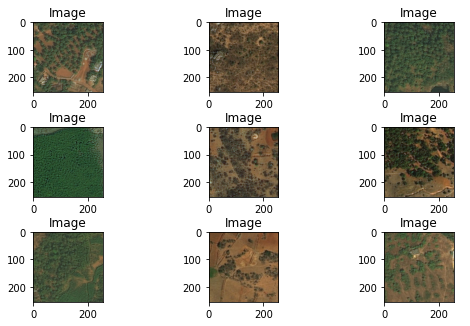}
         \includegraphics[scale = 0.5]{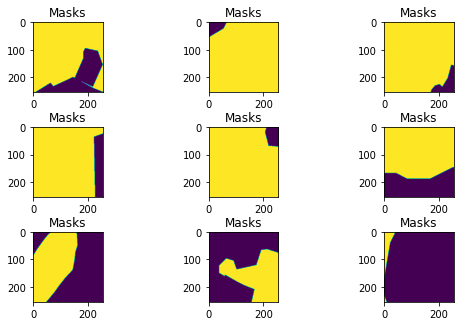}
         \caption{Forest images and their respective masks}
         \label{fig:forest_sample}
\end{figure*}
\begin{figure*}
         \centering
         \includegraphics[scale = 0.5]{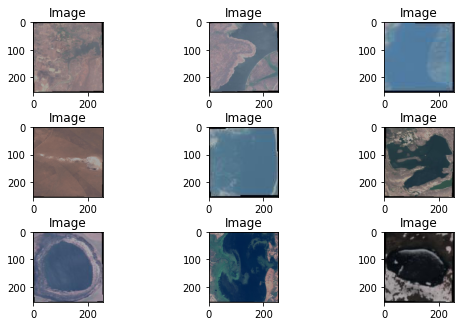}
         \includegraphics[scale = 0.5]{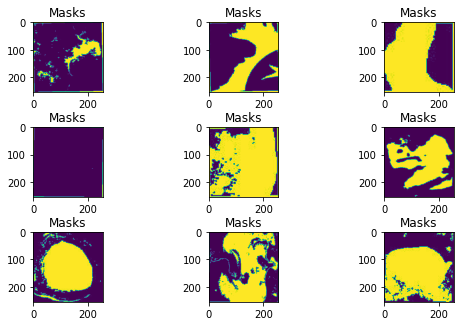}
         \caption{Water body images and their respective masks}
         \label{fig:water_sample}
\end{figure*}

\section{Literature Review}

A lot of work has been done in image segmentation using satellite imaging. Sparavigna \cite{sparavigna2018image} worked on monitoring the water bodies and reservoirs through analysis of micro-graphs using google earth images. They obtained images of the Sarygamysh Lake in Central Asia. Through GIMP, an image processing tool, they obtained the binary mask of the images, and through elevation profile data through google earth, they calculated the amount of water in the lake. Kislov et al. \cite{kislov2021extending} worked on monitoring the damage to the forests induced by bark beetles and windthrows through the use of deep CNN (DCNN). For segmentation of forests, they used the U-Net model. For predicting the forest damage, they obtained accuracy higher than 90\%. They also carried out a comparative analysis between deep CNN and traditional machine learning algorithms, i.e., support vector machine (SVM), random forest, AdaBoost, quadratic discrimination, etc. DCNN outperformed all the other traditional machine learning models. They showed that deep CNN outperformed all the other algorithms. Sampath et al. \cite{sampath2019estimation} proposed a method to estimate the rooftop photovoltaic solar energy generation through satellite imaging. They used U-Net for segmenting the areas available for solar panels. After that, on the basis of incident solar insolation of the area through weather data from the U.S. National Renewable Energy Laboratory (NREL) and energy generation standard values through solar panel datasheets of photovoltaic manufacturers, they calculated the energy generated per year in gigawatt-hour (GWh). Singh et al. \cite{singh2022semantic} used satellite images and deep U-Net for segmentation of different areas in satellite images, i.e., roads, buildings, agricultural lands, etc. They highlighted that this kind of segmentation is essential for sustainable development. To achieve this task, they proposed Deep U-Net, a modified version of U-Net. The main idea behind this segmentation is to find land covers, test their usability, and monitor changes in land covers over time. In the end, they performed a comparative study between SegNet, UNet, and Deep U-Net. According to the evaluation of these models, Deep U-Net outperformed all other models with an Intersection over Union (IoU) value of 89.51 and global accuracy of 90.6\%. Jaisakthi et al. \cite{jaisakthi2021detection} highlighted that analysis of flooded areas is necessary for better response planning. They modified the U-Net architecture to segment the flooded regions through satellite images. In the end, their model gave a 99.46\% IoU value with 99.41\% accuracy. Soni et al. \cite{soni2020m} highlighted that normal U-Net architecture could not extract promising spatial features from satellite images as it is a simplified architecture of image segmentation models. So, they presented a modified version of U-Net that is deep enough to extract contextual information from satellite images. In their modified version, they used DenseNet architecture for downsampling and implemented a long-range skip connection between downsampling and upsampling. Their model achieved an IoU value of 73.02 and 96.02\% overall accuracy. Irwansyah et al. \cite{irwansyah2020semantic} highlighted the problem of city planning in developing countries and proposed a solution for better planning. Their solution includes segmenting the buildings through satellite imaging using U-Net. Their system labeled each pixel in satellite images as building or non-building. They took satellite images of Pasar Minggu Sub-District,
South Jakarta City District, DKI. Jakarta Province. The U-Net model achieved 83\% average training accuracy and 87\% testing accuracy. Alsabhan and Alotaiby \cite{alsabhan2022automatic} highlighted the same problem of city and settlement planning in developing countries. They also worked on building segmentation through satellite imaging. Dataset used by them was an open-source Massachusetts building dataset that was aimed at removing the buildings in the high-density city of Boston. They used U-Net and ResNet-50 models for image segmentation. As a result of experiments, they achieved 82.2\% IoU value and 90\% accuracy for building segmentation. Wang et al. \cite{wang2021semantic} worked on removing the difficulties in early forest fire detection through satellite images. They stated that early forest fires could be mis-detected as clouds. To remove this problem, they worked on the sensitivity of satellite data and remote sensing index from Landsat-8 satellite data. They collected a multispectral high-resolution satellite dataset of forest fires containing different years, seasons, regions, and land cover. They proposed U-Net based architecture named Smoke-Unet. This network consists of an attention mechanism and residual blocks. They showed that RGB, SWIR2, and AOD bands are sensitive to smoke recognition, and their Smoke-Unet showed 3.1\% more accuracy than U-Net. Weng et al. \cite{weng2020water} stated that to study climate change it is necessary to monitor the changes in lakes and rivers. Waterbody segmentation is a crucial part of achieving this task. They highlighted the deficiencies in existing algorithms and proposed their own architecture named separable residual SegNet (SR-SegNet) to achieve this task. Their model surpassed many other states of the art models in segmentation accuracy.

\begin{figure}
    \centering
    \includegraphics[scale=0.5]{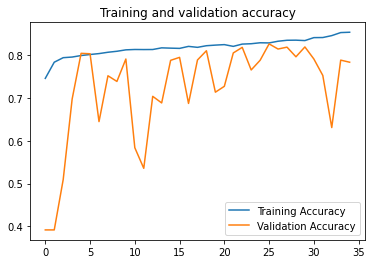}
    \includegraphics[scale=0.5]{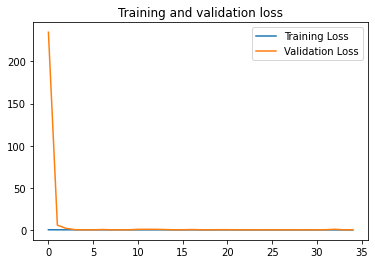}
    \includegraphics[scale=0.5]{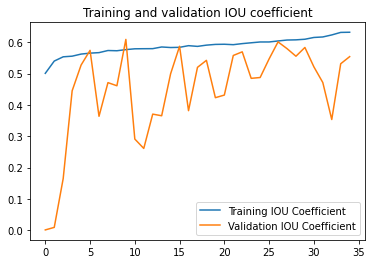}
    \caption{Accuracy, loss and IoU curves for U-Net on forest dataset}
    \label{fig:forest_metric}
\end{figure}

\begin{figure}
    \centering
    \includegraphics[scale=0.5]{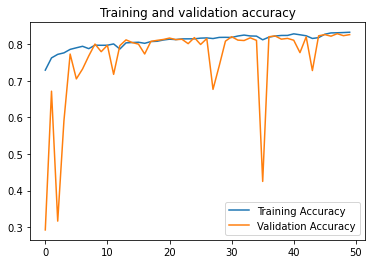}
    \includegraphics[scale=0.5]{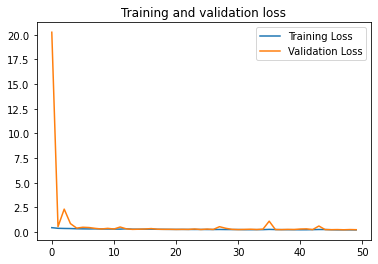}
    \includegraphics[scale=0.5]{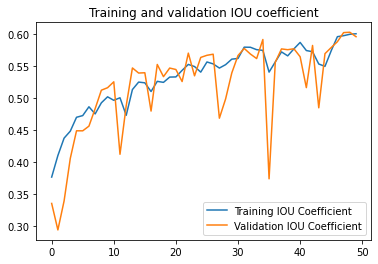}
    \caption{Accuracy, loss and IoU curves for U-Net on water bodies dataset}
    \label{fig:water_metric}
\end{figure}

\section{Methodology}
The overall methodology of implementing the segmentation of forest and lake images was done using U-Net architecture. \autoref{fig:unet} shows the basic architecture of a U-Net. U-Net is based on the principle of encoder and decoder. Just like an encoder and decoder, it has two parts, a contraction part, and an expansion part. The contraction part acts as an encoder and extracts the features through downsampling using max-pooling layers. The expansion part acts as a decoder, and it localizes the segmentation part through transposing convolutional layers. The whole network is end-to-end, fully connected, and does not have any dense layers.

\begin{figure*}
     \begin{subfigure}
         \centering
        \includegraphics[scale=0.6]{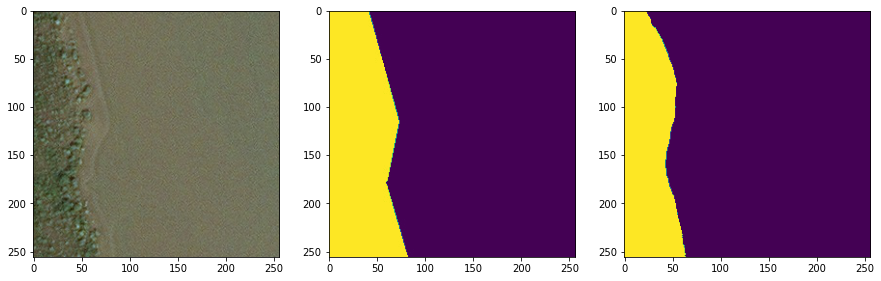}
         \label{fig:y equals x}
     \end{subfigure}

     \begin{subfigure}
         \centering
         \includegraphics[scale=0.6]{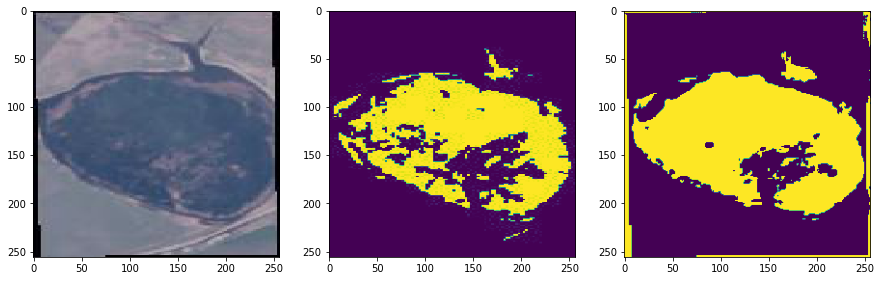}
         \label{fig:three sin x}
     \end{subfigure}
        \caption{Segmentation results for forest (top) and water bodies (bottom) datasets. \textbf{left:} original image, \textbf{center:} ground truth, \textbf{right:} predicted mask }
        \label{fig:img_res}
\end{figure*}

\subsection{Implementation}

The model was trained on a system with 32GB RAM, Ryzon 2500 processor, and Nvidia 1650 4GB graphical processing unit (GPU). Input images given to the U-Net network used were of size 256$\times$256$\times$3. Matrices that were monitored while training was Accuracy, Loss, and Intersection over Union (IoU) coefficient. A learning rate of 0.001 was used along with the binary cross-entropy function. Adam optimizer was used for optimizing the loss function. A batch size of 32 was used for training san for validation batch size of 24 was used. For training 80\% of the dataset was used, and for validation remaining 20\% of the dataset was used. The model was saved whenever minimum loss was achieved. The early stopping criterion was also implemented on the basis of validation loss. If validation loss does not decrease for 9 consecutive iterations, then the model training stops, and the model with the best model weights is used.

\section{Dataset}

Datasets used for training and testing were collected from Kaggle. Forest dataset\footnote{https://www.kaggle.com/datasets/quadeer15sh/augmented-forest-segmentation} was obtained via Land Cover Classification Track in DeepGlobe Challenge \cite{DeepGlobe18}. Dataset consists of 5108 aerial images, where each image has the dimension of 256$\times$256. \autoref{fig:forest_sample} shows some satellite images of forests and their respective masks. Water bodies dataset\footnote{https://www.kaggle.com/datasets/franciscoescobar/satellite-images-of-water-bodies} was captured through Sentinel-2 satellite. Masks of the images were generated by calculating the Normalized Water Difference Index (NWDI). Dataset consists of 2842 images of different dimensions. \autoref{fig:water_sample} shows some satellite images of water bodies and their respective masks

\section{Results \& Discussion}

As explained in the methodology section, U-Net was used to segment the forest and water body images. The model was trained for 50 epochs unless it stopped according to the defined stopping criterion. \autoref{tab:res_table} shows the results obtained for U-Net segmentation model on different datasets. 

\begin{table}[]
    \centering
    \caption{Result matrices for U-Net on forest and water bodies dataset}
    \begin{tabular}{|l|l|l|l|l|}
\hline
Dataset                       & Data Type  & Accuracy & Loss   & IoU    \\ \hline
\multirow{2}{*}{Forest}       & Training   & 0.8277   & 0.37   & 0.6007 \\ \cline{2-5} 
                              & Validation & 0.8255   & 0.3967 & 0.5466 \\ \hline
\multirow{2}{*}{Water bodies} & Training   & 0.8317   & 0.2307 & 0.5974 \\ \cline{2-5} 
                              & Validation & 0.8292   & 0.2385 & 0.6022 \\ \hline
\end{tabular}
    
    \label{tab:res_table}
\end{table}

It can be seen in the table that the U-Net model showed more performance on the water bodies dataset. The reason behind these results is very clear and obvious. The segmentation masks for the forest dataset were generated manually, and many of the masks were not properly labeled. In addition, many masks were not labeled properly according to their corresponding images. While for the water bodies dataset, masks were generated using the system on the basis of NWDI. The learning curved for forest and water bodies datasets are shown in \autoref{fig:forest_metric}, and \autoref{fig:water_metric} respectively.

The first thing to notice from these figures is the number of iterations for which the model was training. U-Net trained for only 35 epochs on the forest dataset; this means that training stopped at epoch 26, and loss didn't change after that. On the other hand, U-Net trained for all 50 epochs, and the accuracy was still increasing. Moreover, the validation accuracy curve of the water bodies dataset is more stable than the forest dataset. Also, if we look at the validation IoU curve for the water bodies dataset, it is still increasing and has an increasing trend from the beginning, while for the forest dataset, the value of the IoU coefficient increased in the beginning and stopped increasing after epoch 5. This shows the importance of a correctly labeled dataset.

In the end, the final results for the segmentation are shown in \autoref{fig:img_res}. The figure shows the original image, their ground truth, and the predicted mask through U-Net. It can be seen here that despite giving the low accuracy, the segmentation results are satisfactory for forest image segmentation.

\subsection{Conclusion \& Future Recommendations}

This paper proposed an application of satellite imaging in forest and water bodies classification using the U-Net segmentation model. Model performance was discussed on the basis of results obtained. The importance of correct data labeling was also highlighted as it caused the decrease in model performance in the forest segmentation dataset.

For future work, it is suggested to firstly remove the mislabeled masks in the forest image segmentation dataset. It is also suggested to use other image segmentation techniques, i.e., DC-UNet or Multi-UNet, etc., to make a comparison with U-Net and select the best model. Moreover, the two datasets can be combined to gather, and a model that can apply segmentation of forest and water bodies separately in a single image can also be implemented. Examples of models that can be used for this task are ResNet50, VGG16, and VGG19.

\bibliographystyle{IEEEtran}
\bibliography{ref}

\begin{thebibliography}{10}
\providecommand{\url}[1]{#1}
\csname url@samestyle\endcsname
\providecommand{\newblock}{\relax}
\providecommand{\bibinfo}[2]{#2}
\providecommand{\BIBentrySTDinterwordspacing}{\spaceskip=0pt\relax}
\providecommand{\BIBentryALTinterwordstretchfactor}{4}
\providecommand{\BIBentryALTinterwordspacing}{\spaceskip=\fontdimen2\font plus
\BIBentryALTinterwordstretchfactor\fontdimen3\font minus
  \fontdimen4\font\relax}
\providecommand{\BIBforeignlanguage}[2]{{%
\expandafter\ifx\csname l@#1\endcsname\relax
\typeout{** WARNING: IEEEtran.bst: No hyphenation pattern has been}%
\typeout{** loaded for the language `#1'. Using the pattern for}%
\typeout{** the default language instead.}%
\else
\language=\csname l@#1\endcsname
\fi
#2}}
\providecommand{\BIBdecl}{\relax}
\BIBdecl

\bibitem{goodfellow2016deep}
I.~Goodfellow, Y.~Bengio, and A.~Courville, \emph{Deep learning}.\hskip 1em
  plus 0.5em minus 0.4em\relax MIT press, 2016.

\bibitem{khryashchev2018comparison}
V.~Khryashchev, L.~Ivanovsky, V.~Pavlov, A.~Ostrovskaya, and A.~Rubtsov,
  ``Comparison of different convolutional neural network architectures for
  satellite image segmentation,'' in \emph{2018 23rd conference of open
  innovations association (FRUCT)}.\hskip 1em plus 0.5em minus 0.4em\relax
  IEEE, 2018, pp. 172--179.

\bibitem{ronneberger2015unet}
O.~Ronneberger, P.~Fischer, and T.~Brox, ``U-net: Convolutional networks for
  biomedical image segmentation,'' in \emph{International Conference on Medical
  image computing and computer-assisted intervention}.\hskip 1em plus 0.5em
  minus 0.4em\relax Springer, 2015, pp. 234--241.

\bibitem{sparavigna2018image}
A.~C. Sparavigna, ``Image segmentation applied to satellite imagery for
  monitoring water in lakes and reservoirs,'' \emph{PHILICA, Article}, no.
  1214, 2018.

\bibitem{kislov2021extending}
D.~E. Kislov, K.~A. Korznikov, J.~Altman, A.~S. Vozmishcheva, and P.~V.
  Krestov, ``Extending deep learning approaches for forest disturbance
  segmentation on very high-resolution satellite images,'' \emph{Remote Sensing
  in Ecology and Conservation}, vol.~7, no.~3, pp. 355--368, 2021.

\bibitem{sampath2019estimation}
A.~Sampath, P.~Bijapur, A.~Karanam, V.~Umadevi, and M.~Parathodiyil,
  ``Estimation of rooftop solar energy generation using satellite image
  segmentation,'' in \emph{2019 IEEE 9th International Conference on Advanced
  Computing (IACC)}.\hskip 1em plus 0.5em minus 0.4em\relax IEEE, 2019, pp.
  38--44.

\bibitem{singh2022semantic}
N.~J. Singh and K.~Nongmeikapam, ``Semantic segmentation of satellite images
  using deep-unet,'' \emph{Arabian Journal for Science and Engineering}, pp.
  1--13, 2022.

\bibitem{jaisakthi2021detection}
S.~Jaisakthi, P.~Dhanya, and S.~Jitesh~Kumar, ``Detection of flooded regions
  from satellite images using modified unet,'' in \emph{International
  Conference on Computational Intelligence in Data Science}.\hskip 1em plus
  0.5em minus 0.4em\relax Springer, 2021, pp. 167--174.

\bibitem{soni2020m}
A.~Soni, R.~Koner, and V.~G.~K. Villuri, ``M-unet: Modified u-net segmentation
  framework with satellite imagery,'' in \emph{Proceedings of the Global AI
  Congress 2019}.\hskip 1em plus 0.5em minus 0.4em\relax Springer, 2020, pp.
  47--59.

\bibitem{irwansyah2020semantic}
E.~Irwansyah, Y.~Heryadi, and A.~A.~S. Gunawan, ``Semantic image segmentation
  for building detection in urban area with aerial photograph image using u-net
  models,'' in \emph{2020 IEEE Asia-Pacific Conference on Geoscience,
  Electronics and Remote Sensing Technology (AGERS)}.\hskip 1em plus 0.5em
  minus 0.4em\relax IEEE, 2020, pp. 48--51.

\bibitem{alsabhan2022automatic}
W.~Alsabhan and T.~Alotaiby, ``Automatic building extraction on satellite
  images using unet and resnet50,'' \emph{Computational Intelligence and
  Neuroscience}, vol. 2022, 2022.

\bibitem{wang2021semantic}
Z.~Wang, P.~Yang, H.~Liang, C.~Zheng, J.~Yin, Y.~Tian, and W.~Cui, ``Semantic
  segmentation and analysis on sensitive parameters of forest fire smoke using
  smoke-unet and landsat-8 imagery,'' \emph{Remote Sensing}, vol.~14, no.~1,
  p.~45, 2021.

\bibitem{weng2020water}
L.~Weng, Y.~Xu, M.~Xia, Y.~Zhang, J.~Liu, and Y.~Xu, ``Water areas segmentation
  from remote sensing images using a separable residual segnet network,''
  \emph{ISPRS International Journal of Geo-Information}, vol.~9, no.~4, p. 256,
  2020.

\bibitem{DeepGlobe18}
I.~Demir, K.~Koperski, D.~Lindenbaum, G.~Pang, J.~Huang, S.~Basu, F.~Hughes,
  D.~Tuia, and R.~Raskar, ``Deepglobe 2018: A challenge to parse the earth
  through satellite images,'' in \emph{The IEEE Conference on Computer Vision
  and Pattern Recognition (CVPR) Workshops}, June 2018.

\end{thebibliography}

\end{document}